%% file: main.tex
\documentclass[conference]{IEEEtran}
\newcommand{\nop}[1]{}
\usepackage{algorithm}
\usepackage{algpseudocode}
\usepackage{subfigure}
\usepackage{graphicx}
\usepackage{multirow}
\usepackage{amsmath,amssymb,amsfonts}
\usepackage{bm}
\usepackage{booktabs}
\usepackage{hyperref}
\usepackage{algorithm}
\usepackage{multirow}
\usepackage{amsmath,amssymb,amsfonts}
\usepackage{amsthm}
\usepackage{graphicx}
\usepackage{setspace}
\usepackage{multicol}
\usepackage{multirow}
\usepackage{cite}
\usepackage{float}
\usepackage{color}
\usepackage{alltt, listings}
\usepackage{subfigure}
\usepackage{algpseudocode}
\usepackage{booktabs}
\usepackage{balance}
\usepackage{wrapfig}
\usepackage{url}
\usepackage{bm}
\usepackage[textsize=tiny]{todonotes}
\usepackage[utf8]{inputenc} 
\usepackage[T1]{fontenc}
\usepackage{url} 
\usepackage[left=1.62cm,right=1.62cm,top=1.9cm]{geometry}
\newtheorem{assumption}{Assumption}
\newtheorem{theorem}{Theorem}
\newtheorem{lemma}{Lemma}

\newtheorem{proposition}{Proposition}
\newtheorem{remark}{Remark}
\theoremstyle{definition}
\newtheorem{definition}{Definition}

\begin{document}

\title{Killing Two Birds with One Stone: Quantization Achieves Privacy in Distributed Learning}

\author{\IEEEauthorblockN{Guangfeng Yan$^{1}$\quad Tan Li$^{1}$ \quad Kui Wu$^{2}$\quad Linqi Song$^{1}$}
\IEEEauthorblockA{
$^{1}$ Department of Computer Science, City University of Hong Kong\\
$^{2}$ Department of Computer Science, University of Victoria}}

\maketitle

\begin{abstract}
Communication efficiency and privacy protection are two critical issues in distributed machine learning. Existing methods tackle these two issues separately and may have a high implementation complexity that constrains their application in a resource-limited environment. We propose a comprehensive quantization-based solution that could simultaneously achieve communication efficiency and privacy protection, providing new insights into the correlated nature of communication and privacy. Specifically, we demonstrate the effectiveness of our proposed solutions in the distributed stochastic gradient descent (SGD) framework by adding binomial noise to the uniformly quantized gradients to reach the desired differential privacy level but with a minor sacrifice in communication efficiency. We theoretically capture the new trade-offs between communication, privacy, and learning performance. 
\end{abstract}
\input{Key_Word}
\input{Introduction}
\input{Related_Work}
\input{Problem_Formulation}

\input{BQ_scheme}

\input{Experiments}

\input{Conclusion}
\bibliographystyle{IEEEtran}
\bibliography{mybib}
\newpage
\input{Appendix}

\end{document}

%% file: Key_Word.tex
\begin{IEEEkeywords}
Distributed Learning, Communication Efficiency, Quantization, Privacy
\end{IEEEkeywords}

%% file: Introduction.tex
\section{Introduction}
\label{introduction}

Distributed machine learning has recently attracted more attention due to the distributed ways of collecting and processing data, such as federated learning. Communication efficiency and privacy protection are two critical concerns in such an emerging paradigm. Various deep learning models need to be exchanged among distributed computing nodes, which are often subject to the communication bottleneck \cite{tang2020communication,tao2018esgd}. Furthermore, exchanging these model parameters or even just gradients may lead to the privacy leakage of local data~\cite{zhu2020deep,fredrikson2015model,nasr2019comprehensive}.

Existing works proposed separate methods to tackle these two issues and made corresponding progress. Communication reduction can be achieved by some compression techniques ~\cite{alistarh2017qsgd, basu2019qsparse, nadiradze2021asynchronous, mcmahan2017communication} while privacy leakage can be constrained to a certain level by add-on noise to disturb the raw updated gradients~\cite{dwork2014algorithmic, bonawitz2017practical}. While some works~\cite{zong2021communication,mohammadi2021differential} try to tackle the two problems together, they simply combine the above techniques, e.g., compressing the perturbed gradient, without offering a theoretical guarantee on the final model performance. Besides, none of them consider the privacy proprieties of compression. We argue that the above two operations are not completely independent. In particular, quantization, one of the representative compression techniques, has inherited privacy properties. Therefore, we can kill two birds with one stone.

In this paper, we propose a new quantization-based solution to achieve communication efficiency and privacy protection simultaneously in a distributed stochastic gradient descent (SGD) framework. Instead of performing extra processing on add-on noise, our solution utilizes the inherent quantization noise to achieve the desired level of privacy. Specifically, our proposed \textbf{B}inomial mechanism aided \textbf{Q}uantized (BQ) scheme adds appropriately parameterized binomial distributed noise to the quantized gradient, which can be combined together into a unified quantization process, to reach the required privacy with only a slight sacrifice in communication efficiency. We theoretically analyze the privacy leakage of the quantized gradients in a differential privacy (DP) manner: how does removing one sample in the training dataset impact the output of the quantization mechanism?

Using the inherent privacy properties of BQ scheme, we design a novel binomial mechanism-aided quantized SGD (BQ-SGD) algorithm, which can satisfy the required privacy communication budget by adjusting both the quantization level and binomial noise level. We theoretically characterize the trade-offs between communication, privacy, and learning performance under this quantization scheme.

%% file: Related_Work.tex
\section{Related Work}
Several schemes have been proposed to improve the communication efficiency in gradient-based large-scale distributed learning: sparsification~\cite{shi2019distributed}, sketching~\cite{rothchild2020fetchsgd}, quantization~\cite{alistarh2017qsgd}, less frequent communication~\cite{mcmahan2017communication,zhang2020lagc}, or their combinations~\cite{basu2019qsparse,nadiradze2021asynchronous}. In particular, we pay special interest to quantization, which is the basic technique that changes the floating point numbers into fixed point ones. In contrast, other techniques can be used together with quantization in parallel. 

Privacy preservation for machine learning has been studied in distributed learning algorithms to prevent privacy leakage from the model exchange. 
To counter this issue, most existing efforts~\cite{abadi2016deep} applied DP mechanism to add controllable noise to the 
to raw gradients. The trade-off between learning convergence and privacy has been  theoretically studied in ~\cite{wei2020federated}.  

Some recent work has considered communication-efficiency and privacy-preserving abilities together in distributed learning~\cite{zong2021communication,mohammadi2021differential,agarwal2018cpsgd}, however, in a straightforward way via a combination of adding noise and quantizing the noisy gradient. Different from them, the privacy properties of compression have been observed by some early works~\cite{xiong2016randomized, jonkman2018quantisation}, which illustrate that one can translate the perturbation incurred by the compression method into measurable noise to ensure privacy. However, none of them built a theoretical framework to understand this or theoretically proves the privacy guarantee that quantification itself can provide.

%% file: Problem_Formulation.tex
\section{Problem Formulation}
We consider a distributed learning problem, where $N$ clients collaboratively train a shared model via a central parameter server. The local dataset at client $i$ is denoted as $D^{(i)}$. Our goal is to find a set of global optimal model parameters $\theta$ by minimizing the objective function $F: \mathbb{R}^d \to \mathbb{R}$,
\begin{equation}
\begin{array}{llll}	\min_{\theta \in \mathbb{R}^d}F(\theta) = \sum_{i=1}^N  p_i \mathbb{E}_{\xi^{(i)}\sim D^{(i)}}[l(\theta;\xi^{(i)})],
	\end{array}\label{optim_problem}
\end{equation}
where 
$p_i$ \footnote{ Note that this work focuses on exploring the impact of server-client communication, rather than the specific server-side aggregation on learning performance. We claim that our proposed method can be used in combination with different aggregation methods, such as \cite{mcmahan2017communication} ~\cite{blanchard2017machine} or~\cite{yin2018byzantine}.} is the weight of client $i$; $l(\theta;\xi^{(i)})$ is the local loss function of the model $\theta$ towards one data sample $\xi^{(i)}$; and the expectation is taken with respect to the sampling randomness of data $\xi^{(i)}$.

A standard approach to solve this problem is SGD and its variations, such as momentum, or Adam. Without loss of generality, we describe our framework with the classic setting with one local update (one step of gradient descent) for each iteration. Note that our method also works for these gradient-based variations. At iteration $t$, each client $i$ first downloads the global model $\theta_t$ from server, then randomly selects a batch of samples $B^{(i)}_t \subseteq D^{(i)}$ with size $L$ to compute its local stochastic gradient with model parameter $\theta_t$:
\begin{equation}
\begin{array}{llll}
\mathbf{g}^{(i)}_t = \mathcal{G}(D^{(i)}) := \frac{1}{L} \sum_{\xi^{(i)} \in B^{(i)}_t: B^{(i)}_t \subseteq D^{(i)}} \nabla l(\theta_t;\xi^{(i)}). \end{array}
\label{eq:sgrad} 
\end{equation}Then the server aggregates these gradients and sends the updated model $\theta_{t+1}$ back to all clients:
\begin{equation}
\begin{array}{llll}
    \theta_{t+1} = \theta_t - \eta\sum_{i=1}^N p_i\mathbf{g}^{(i)}_t
    \end{array}
\end{equation}
where $\eta$ is the server learning rate. We make the following two common assumptions on the raw gradient $\mathbf{g}_t^{(i)}$ and the objective function $F(\theta)$~\cite{{bottou2018optimization},{tang2019doublesqueeze}}: 
\begin{assumption}[Unbiasness and Bounded Variance of $\mathbf{g}_t^{(i)}$]
	The stochastic gradient oracle gives us an independent unbiased estimate $\mathbf{g}_t^{(i)}$ with a bounded variance:
	\begin{align}
	\mathbb{E}_{B_t^{(i)}}[\mathbf{g}^{(i)}_t] = \nabla F(\theta_t),~	\mathbb{E}_{B_t^{(i)}}[\|\mathbf{g}^{(i)}_t-\nabla F(\theta_t)\|^2] \le \frac{\sigma^2}{L}.
	\end{align}
	\label{ass:stochastic_gradient} 
\end{assumption}
\vspace{-5mm}
\begin{assumption}[Smoothness]
	The objective function $F(\theta)$ is $\nu$-smooth: $\forall \theta,\theta' \in \mathbb{R}^d$, $\|\nabla F(\theta)-\nabla F(\theta')\| \leq \nu\|\theta-\theta'\|$.
	\label{ass:smoothnesee} 
\end{assumption}
Assumption~\ref{ass:smoothnesee} further implies that $\forall \theta,\theta' \in \mathbb{R}^d$, we have \begin{equation}
F(\theta') \leq F(\theta) + \nabla F(\theta)^\mathrm{T} (\theta'-\theta) + \frac{\nu}{2} \|\theta'-\theta\|^2.
\label{eq:smooth_1}
\end{equation}


In vanilla distributed SGD, client $i$ directly transmits the raw stochastic gradient $\mathbf{g}^{(i)}_t$ to the server, which may pose a significant communication burden and privacy risk. To cope with the two drawbacks, we propose quantization schemes that convert $\mathbf{g}_t^{(i)}$ to a perturbed version and carefully characterize the inherent privacy guarantee it can provide using a DP-like definition. Based on the proposed quantization scheme, we present a new variant of distributed SGD that can achieve the desired level of privacy and communication constraints. 



%% file: BQ_scheme.tex
\section{Binomial Mechanism Aided Quantization} \label{sec:BMQ-SGD} In this section, we propose a three-step quantization mechanism, called Binomial mechanism aided Quantization (BQ) and carefully characterize its inherent privacy guarantee.  
\subsection{BQ scheme}
\textbf{Step 1: Norm Clipping} We first clip the gradient of each sample $\nabla l(\theta_t;\xi^{(i)})$ of Eq.~(\ref{eq:sgrad}) into $l_{\infty}$ norm with threshold $C$: 
\begin{equation}
	\nabla \hat{l}(\theta_t;\xi^{(i)}) = \frac{\nabla l(\theta_t;\xi^{(i)})}{\max{\{1, \|\nabla l(\theta_t;\xi^{(i)})\|_{\infty}/C\}}}. \label{eq:clipoperation} 
\end{equation}
We then calculate the clipped gradient by averaging all $\nabla \hat{l}(\theta_t;\xi^{(i)})$ in a batch :\begin{equation}
	\begin{array}{llll}
		\mathbf{g}^{(i),C}_t =  \frac{1}{L} \sum_{\xi^{(i)} \in B^{(i)}_t}\nabla\hat{l}(\theta_t;\xi^{(i)}).\end{array}\label{eq:clipgrad} 
\end{equation}
Since $\|\nabla \hat{l}(\theta_t;\xi^{(i)})\|_{\infty} \le C$,  referring to the triangular inequality of $l_{\infty}$ norm, we have $\|\mathbf{g}^{(i),C}_t \|_{\infty} \le C$. The clipped gradient here limits the impact of a single sample on the whole, and it also plays a role in the privacy analysis. 

\textbf{Step 2: Uniform Quantization} We then introduce a commonly used quantization scheme, namely, uniform quantization~\cite{alistarh2017qsgd} to quantize the clipped gradients in an element-wise way. In particular, the $j$-th element of $\mathbf{g}^{(i),C}_t$, denoted as $g_j$ for simplicity, is quantized as
\begin{equation}\
	\mathcal{Q}_b[g_j] = C \cdot \text{sgn}(g_j)\cdot\rho(g_j,s),\label{quantizedoperation}
\end{equation} 
where $\mathcal{Q}_b[\cdot] $ is a $b$-bit quantizer, $\text{sgn}(g_j)=\{+1,-1\}$ is the sign of $g_j$, $s$ is the given quantization level, and $\rho(g_j,s)$ is an unbiased stochastic function that maps scalar $|g_j|/C$ to one of the values in set $\{0, 1/s, 2/s, \ldots, s/s\}$. For example, for $|g_j|/C \in [l/s,(l+1)/s]$, we have
\begin{equation}
	\rho(g_j,s) = \begin{cases} l/s,& \text{w.p. $1-p$,}\\ 
		(l+1)/s,& \text{w.p. $p = \frac{s|g_j|}{C}-l$.} \end{cases}
	\label{eq:map}
\end{equation} 
 The quantization level $s$ is roughly exponential to the number of quantized bits $b$. If we use one bit to represent its sign and the other $b-1$ bits to express $\rho(g_j,s)$, we can achieve quantization level $s=2^{b-1}-1$. 
After quantization, each client obtains a tuple $(\bm{\sigma}^{(i)}_t, \bm{\rho}^{(i)}_t)$, where $\bm{\sigma}^{(i)}_t$ and $\bm{\rho}^{(i)}_t$ are the vectors of signs and integer values (i.e., $s\cdot\rho(g_j,s)$) for all elements of $\mathbf{g}^{(i),C}_t$. We define the gradient after uniform quantization as: 
\begin{equation}
	\begin{array}{llll}
\mathbf{g}_t^{(i),U} = \frac{C}{s}\cdot \bm{\sigma}^{(i)}_t \odot \bm{\rho}^{(i)}_t,\label{eq:UQ}
\end{array} 
\end{equation}
where $\odot$ denotes the element-wise product. 

We next show that uniform quantization can provide privacy protection in some cases. In particular, we study the privacy loss of the $\mathbf{g}_t^{(i),U}$ in a DP manner: how does removing or adding one sample in the original dataset impact the output gradient. The formal definition of $(\epsilon, \delta)$-DP for SGD is as follows.
\begin{definition}[$(\epsilon, \delta)$ - DP \textbf{for SGD}]
	Given a set of data sets $\mathcal{D}$ and a query function $q: \mathcal{D} \to \mathcal{X}$, a mechanism $\mathcal{M}: \mathcal{X} \to \mathcal{O}$ to release the answer of the query, is defined to be $(\epsilon, \delta)$ - DP if for any adjacent datasets $(D, D') \in \mathcal{D} \times \mathcal{D}$ and any measurable subset outputs $O \in \mathcal{O}$,
	\begin{equation}
		\Pr\{\mathcal{M}[q(D)]\in O\}\leq \Pr\{\mathcal{M}[q(D')]\in O\}e^{\epsilon} + \delta, \label{eq:DP}
	\end{equation}
	\label{def:DP}
\end{definition}
\vspace{-20pt}
\noindent where $\epsilon>0$ is the distinguishable bound of all outputs on adjacent datasets $D,D'$ that differ in at most one data sample. $\delta$ represents the event that the ratio of the probabilities for two adjacent datasets $D,D'$ cannot be bounded by $e^\epsilon$ after privacy-preserving mechanism $\mathcal{M}$. In the rest of the paper, we also call a solution \textit{SGD private} if it meets the above definition.

In the SGD framework, $\mathcal{G}(D)$ is viewed as a \textit{query function} that maps dataset $D$ to a \textit{query result} (i.e., raw gradient $\mathbf{g}$). 
The SGD privacy ensures that any element of the $\mathbf{g}$ is “essentially” equally likely to occur, independent of the presence or absence of any individual data sample. We denote $\mathbf{g},\mathbf{g}'$ as two raw gradients computed based on two adjacent datasets $D,D'$, and $g, g'$ as two elements in the same entry of $\mathbf{g}, \mathbf{g}'$. By considering uniform quantization as a mechanism $\mathcal{M}$ that maps query results (raw gradients $\mathbf{g}$) to the perturbed outputs (quantized gradient $\mathbf{g}^{U}$), our criterion for detecting whether uniform quantization satisfies SGD privacy is: whether the original $g$ and $g'$ can be distinguished through the disturbed $g^U$ and $g^{'U}$.

\vspace{-1pt}
\begin{figure}[htbp]
	\centering
	\subfigure[Uniform Quantization ($s=10$)]{
		\includegraphics[width=0.8\columnwidth]{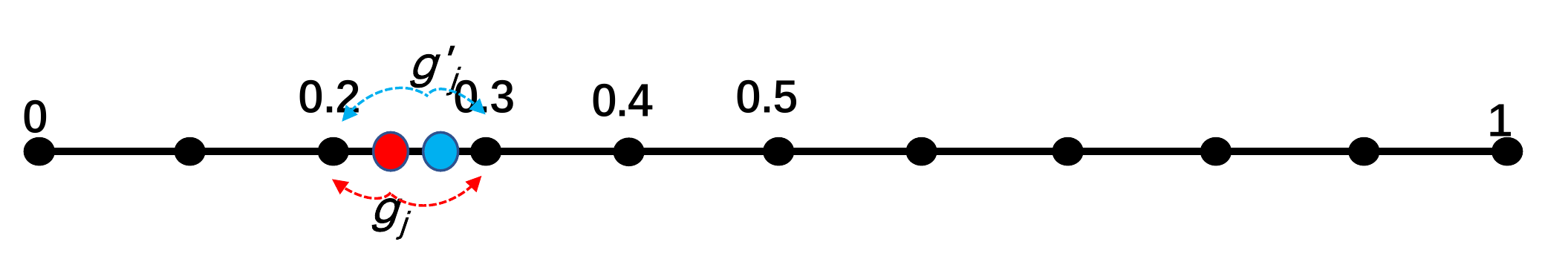}\label{fig:qsgd1}
	}
	\subfigure[Uniform Quantization ($s=10$)]{
		\includegraphics[width=0.8\columnwidth]{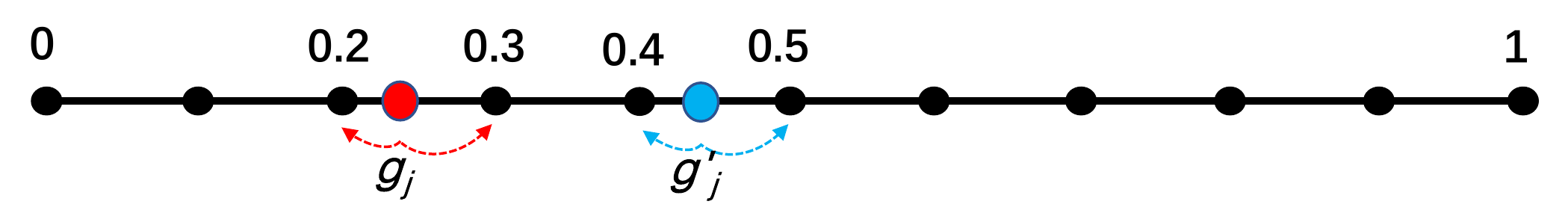}\label{fig:qsgd2}
	}
	\subfigure[Binomial Mechanism Aided Quantization ($s=10$,  $m=1$)]{
		\includegraphics[width=0.8\columnwidth]{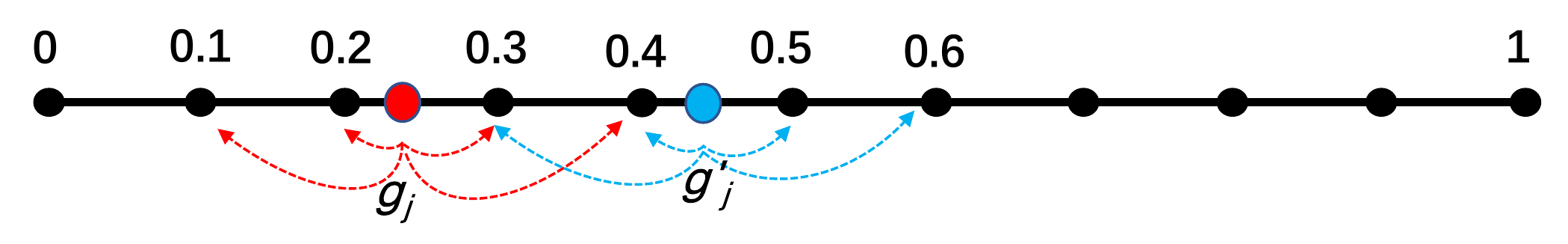}\label{fig:mqsgd}
	}
	\caption{An illustration of quantized gradients after step 2 and step 3. }
	\label{fig.qsgd}
\end{figure}

The uniform quantization maps the raw value to the two nearest quantization points, see Eq.~\eqref{eq:map} and Fig.~\ref{fig.qsgd}. In Fig.~\ref{fig:qsgd1}, $g$ and $g'$ are within interval $[0.2,0.3]$. We say the uniform quantization satisfies SGD privacy if $g$ and $g'$ are mapped to the same quantization points, i.e. $g^U= g^{'U} =0.2 (\text{or }0.3)$. Fig.~\ref{fig:qsgd2} shows the case that uniform quantization fails to provide SGD privacy. $g$ and $g'$ fall in different intervals. After uniform quantization, $g^U=0.2$ or $0.3$, while $g^{'U}=0.4$ or $0.5$, which $g^U$ and $g^{'U}$ can always be distinguished. This example suggests that \textbf{Step 1 and Step 2 together cannot achieve SGD privacy.}

\textbf{Step 3: Binomial Noise Addition} To enhance the SGD privacy guarantee, additional noise should be added after step 2. Inspired by~\cite{agarwal2018cpsgd}, we add an extra noise $\mathbf{o}^{(i)}_t$ on the values after uniform quantization, which are i.i.d sampled from a Binomial distribution $Bin(m,q)$, to enhance the privacy guarantee with only a slight sacrifice in communication. Fig.~\ref{fig:mqsgd} shows that the binomial noise addition allows $g^U$ and $g^{'U}$ to reach the yonder points, like $[0.3, 0.4]$, thus achieving SGD privacy. Formally, the uniform Quantization maps $g_j$ into the values in set $\{-s, 1-s, ..., 0, 1, ..., s-1, s\}$. After binomial noise addition, the set is extended to $\{-s, 1-s, ..., 0, 1, ..., s+m-1, s+m \}$, which includes $2s+m+1$ points. That is, $\log(2s+m+1)$ bits are needed for transmission. The client $i$ sends $\bm{\sigma}_t^{(i)} \odot \bm{\rho}_t^{(i)} + \bm{o}_t^{(i)}$ instead of raw gradient $\mathbf{g}_t^{(i)}$ to the server. To ensure the unbiased estimation of $\mathbf{g}_t^{(i)}$, the server decodes the tuple and computes the BQ gradient as: 
\begin{equation}
	\begin{array}{llll}
		\mathbf{g}_t^{(i),B} = \mathcal{G}_B(D^{(i)})=  \frac{C}{s}\cdot [\bm{\sigma}_t^{(i)} \odot \bm{\rho}_t^{(i)} + \mathbf{o}_t^{(i)} - mq\cdot\bm{1}],
	\end{array}
	\label{eq:BQ} 
\end{equation}
where $\bm{1}$ is an all-one vector. 

\subsection{Inherit privacy guarantee of BQ scheme}
\begin{figure}[ht]
\vskip -0.05in
\begin{center}
\centerline{\includegraphics[width=0.7\linewidth]{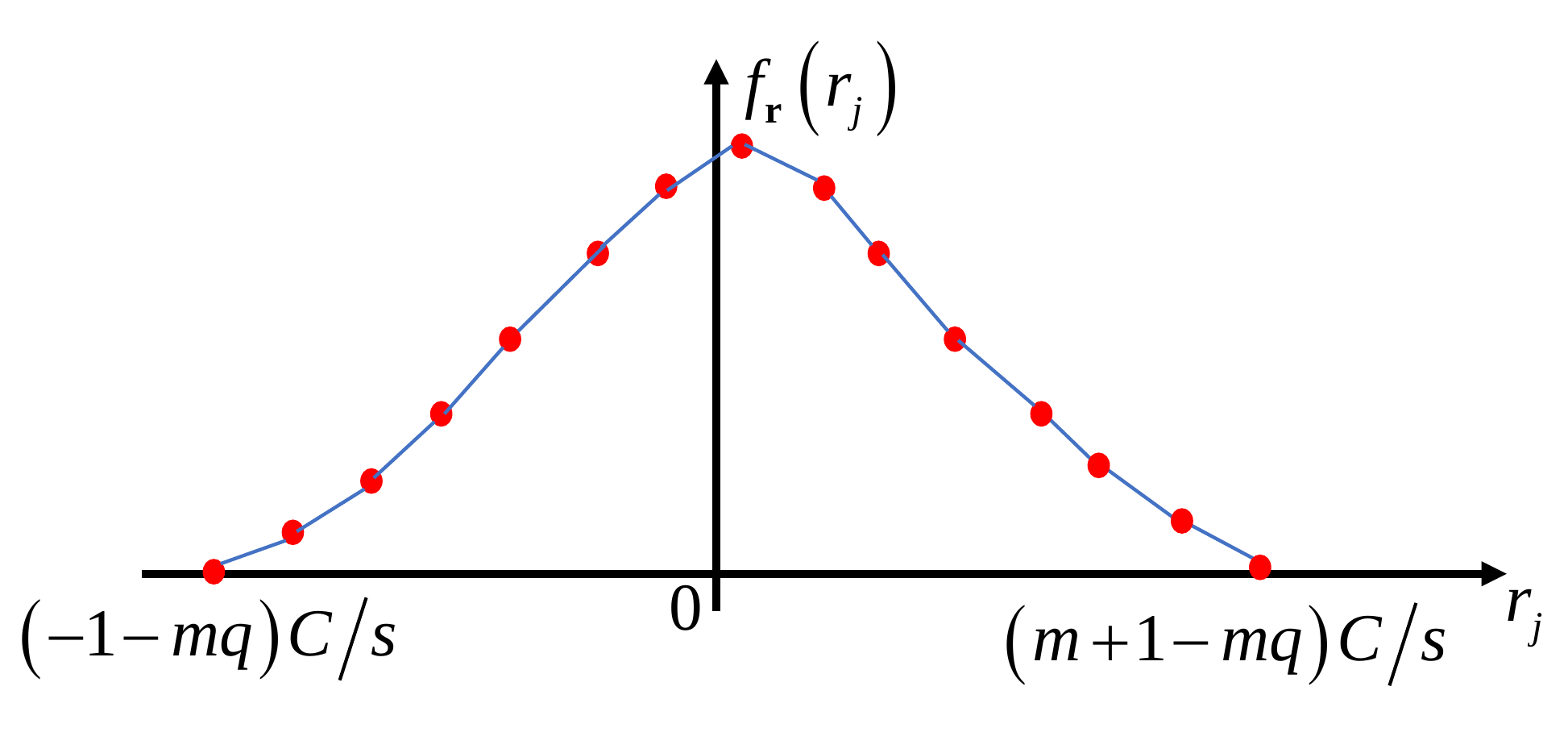}}
	\caption{Noise shape of BQ scheme.}
	\label{fig:noise}
	\vspace{-0.15in}
\end{center}
\vskip -0.2in
\end{figure}

We then investigate how the BQ gradient $\mathbf{g}_t^{(i),B}$ protects privacy of dataset $D^{(i)}$ being used for model training. Most DP-manner mechanisms, like Laplace or Gaussian mechanism, add controlled noise from some predetermined distributions on raw gradient $\mathbf{g}$ to ensure privacy. Analogously, the BQ gradient $\mathbf{g}_t^{(i),B}$ can be viewed as the noisy version of original one $\mathbf{g}_t^{(i)}$, namely, $\mathbf{g}_t^{(i),B} = \mathbf{g}_t^{(i)} + \mathbf{r}_t^{(i)}$. Therefore, the properties of BQ noise $\mathbf{r}_t^{(i)}$ is the key factor for the privacy analysis. We show the properties of the quantization noise in Proposition~\ref{prop:1}.
\begin{proposition}[Properties of BQ noise]	\label{prop:1}
	After performing $C$-norm clipping, $s$-level quantization and adding binomial noise vector i.i.d sampled from $Bin(m,q)$, the probability density of the $j$-th element of $\bm{r}_t^{(i)}$, denote as $r_j$, is: 
	\begin{equation}
		\begin{array}{llll}
		&f_{\bm{r}}(r_j) = \frac{s}{C}[(k+1-mq)P_k + (mq-k)P_{k+1}] \\
		&~~~~~~~~~~~+ \frac{s^2}{C^2}(P_{k+1}-P_k) r_j,\label{eq:noise} \end{array}
	\end{equation}
 \vspace{-10pt}
 \begin{align}
		\begin{array}{llll}
			 r_j \in [\frac{(k-mq)C}{s}, \frac{(k+1-mq)C}{s}], k=-1,0,...,m.\nonumber
		\end{array}
	\end{align}  
	where $P_k = \binom{m}{k} q^k(1-q)^{m-k}$ , $P_{-1} = P_{m+1} = 0$. 
	
	\noindent The statistic properties of BQ noise are : $\mathbb{E}[\bm{r}] = \bm{0}$ and 
	\begin{align}
		\begin{array}{llll}
			\mathbb{E}[\|\bm{r}\|^2] = dC^2 \Big[\frac{mq(1-q)}{s^2}+\frac{1}{6s^2}\Big] \triangleq dC^2V(m, q, s)\label{eq:BQvariance}
		\end{array}
	\end{align} 
\end{proposition}
We define $V(m, q, s)\triangleq \frac{mq(1-q)}{s^2}+\frac{1}{6s^2}$ as the BQ noise variance. The proof of Proposition \ref{prop:1} is in Appendix~\ref{pro:Pro1}. The noise shape of BQ scheme is shown in Fig.~\ref{fig:noise}. Compared to CQSGD \cite{agarwal2018cpsgd}, which only considers Binomial Noise Addition (e.g., Step 3), we comprehensively consider the noise generated by Uniform Quantization and Binomial Noise Addition (e.g., Step 2 and Step 3). Therefore, the noise of BQ scheme is a continuous variable relative to the binomial noise in CQSGD, and the noise shape is close to Gaussian noise.

Using Proposition~\ref{prop:1}, we can derive the privacy guarantee provided by BQ quantized gradient as follows. 
\begin{lemma}[Privacy guarantee of BQ Gradient]\label{lemma1}
	The BQ gradient $\tilde{\mathbf{g}}_t^{(i),B}$ satisfies $(\epsilon^{(i)}, \delta^{(i)})$-SGD privacy with $\epsilon^{(i)} = \frac{8dsLP_{max}}{|D^{(i)}|^2 \delta^{(i)}}$, where $P_{max} = max_k\{P_k\}$.
\end{lemma}
The proof of Lemma \ref{lemma1} is in Appendix~\ref{pro:lemma1}.

\begin{remark}
	For general $m$, $q=\frac{1}{2}$ is the optimal choice as it can minimize $P_{max}$. According to De Moivre–Laplace theorem, for $m>10$, $P_{max}$ can be expressed as
	\begin{equation}
		\begin{array}{llll}
			P_{max} = \max_k\{\frac{1}{\sqrt{2\pi\sigma^2}}\exp{\frac{-(k-\mu)^2}{2\sigma^2}}\}=\frac{1}{\sqrt{2\pi\sigma^2}}.
		\end{array}
	\end{equation}
	where $\sigma^2=\frac{m}{4}$. Hence, $\epsilon^{(i)} = \frac{6.4dsL}{|D^{(i)}|^2\sqrt{m} \delta^{(i)}}$.
\end{remark}

\begin{remark}
The privacy provided by the BQ gradient is affected by three key factors: (i) the dimension of gradient $d$, since each dimension may reveal privacy; (ii) the ratio of batch sampling $L/|D^{(i)}|$. The small ratio leads to the small $\epsilon$ (i.e. stronger privacy) according to the privacy amplification theorem~\cite{balle2018privacy}; (iii) quantization level $s$ and binomial noise $m$, the lower quantization level $s$ and larger binomial noise $m$ contribute to stronger privacy protection. Noting since $s$ is always larger than 1 and $2s+m+1<2^b$, to provide $(\epsilon, \delta)$-SGD privacy, the required bits $b$ for transmission should be no less than $\frac{1}{2}\log \frac{6.4dL}{|D^{(i)}|^2} - \frac{1}{2}\log \epsilon \delta$.
\end{remark}
\vspace{-10pt}

\subsection{Algorithm Description}
The inherent privacy properties of the BQ scheme are stated in Lemma~\ref{lemma1}. However, many real-world applications will give the privacy requirement and communication budget in advance by considering the sensitivity of local datasets, channel conditions, and tolerant latency. In this section, we propose the following algorithm, Binomial Mechanism aided Quantized distributed SGD (BQ-SGD), where each client takes explicit communication constraint $\bar{b}^{(i)}$ and privacy requirement.
\begin{algorithm}[ht] 
	\caption{Binomial Mechanism aided Quantized distributed SGD (BQ-SGD) } 
	\begin{algorithmic}[1] 
		\State \textbf{Input:} Learning rate $\eta$, initial point $\theta_0 \in \mathbb{R}^d$, gradient norm bound $C$; set communication constraint $\bar{b}^{(i)}$ and privacy requirement $(\bar{\epsilon}^{(i)}$, $\bar{\delta}^{(i)})$ for client $i$.
		\For {each iteration $t = 0, 1, ..., T-1$:}
		\State \textbf{On each client {$i=1, ..., N$}:}
		\State Receive $\theta_t$ from server;
		\State Compute $\nabla l(\theta_t;\xi^{(i)})$ for each sample  $\xi^{(i)} \in B^{(i)}_t$;
		\State Clipping $\nabla l(\theta_t;\xi^{(i)})$ to $\nabla\hat{l}(\theta_t;\xi^{(i)})$ using Eq.~(\ref{eq:clipoperation});
         \State Determine $(s^{(i)}, m^{(i)})$ for client $i$ using Eqs.~(\ref{eq:s}) (\ref{eq:m});
		\State Compute parameters $\bm{\sigma}_t^{(i)}, \bm{\rho}_t^{(i)}$ for uniform quantizer, and $\bm{o}_t^{(i)}$, where $\mathbf{o}_t^{(i)}$ are i.i.d. samples from $Bin(m^{(i)},\frac{1}{2})$;
		\State Send $(\bm{\sigma}_t^{(i)} \odot  \bm{\rho}_t^{(i)}+ \bm{o}_t^{(i)})$ to the server;
		\State \textbf{On the server:}
		\State Decode ${\mathbf{g}}_t^{(i),B}$ according to Eq.~(\ref{eq:BQ});
		\State Aggregate ${\mathbf{\bar g}}_t \triangleq \sum_{i=1}^N p_i{\mathbf{g}}_t^{(i),B}$;
		\State Update model parameter: $\theta_{t+1} = \theta_t - \eta\mathbf{\bar g}_t$;
		\State Send  $\theta_{t+1}$ to all clients;
		\EndFor
	\end{algorithmic} 
	\label{alg:BQSGD}
\end{algorithm}

The key step of Alg.~\ref{alg:BQSGD} is Line 7, which determines the quantization level $s^{(i)}$ and binomial noise level $m^{(i)}$ to meet the needs of $\bar{b}^{(i)}$ and $(\bar{\epsilon}^{(i)}, \bar{\delta}^{(i)})$ for each client. Note that if $\bar{b}^{(i)}$ and $(\bar{\epsilon}^{(i)}, \bar{\delta}^{(i)})$ are invariant during $T$ iterations, we only need to calculate $s^{(i)}$ and $m^{(i)}$ once before training starts. But if the requirements change, we need to recalculate the parameters for the BQ scheme. For simplicity, we omit $i$ and $t$ in the following analysis. We seek a pair of $(s, m)$ that minimize the convergence error under privacy and communication constraints, which is further equivalent to minimizing the BQ noise variance $V(m,q,s)$ defined in Eq.~\eqref{eq:BQvariance}. By setting $q = \frac{1}{2}$, we can formally formulate this constrained optimization problem as:
\begin{align}
	&\min_{m,s} V(m, \frac{1}{2},s) = \frac{m}{4s^2}+\frac{1}{6s^2} \\
	&s.t ~~\log(2s + m + 1) \le \bar{b},\label{eq:cconstraint}\\
	&~~~~~~~\frac{6.4 dsL}{|D|^2 \sqrt{m}\bar{\delta}} = \bar{\epsilon}.\label{eq:pconstraint}
\end{align}
Eq.~\eqref{eq:cconstraint} captures the communication constraint using $2s+m< 2^{\bar{b}}-1$. Eq.~\eqref{eq:pconstraint} indicates the privacy constraint, which can be directly derived from Lemma~\ref{lemma1}. By solving the above optimization problem, we have,
\begin{align}
	&	s = \bar{R}\sqrt{{\bar{R}}^2+(2^{\bar{b}}-1)}-{\bar{R}}^2,\label{eq:s}\\
	&	m = \frac{s^2}{{\bar{R}}^2} = (2^{\bar{b}}-1)+2{\bar{R}}^2-2\bar{R}\sqrt{{\bar{R}}^2+(2^{\bar{b}}-1)}\label{eq:m}
\end{align}
where $\bar{R}=\frac{\bar{\delta}\bar{\epsilon}|D|^2 }{6.4dL}$ is proportional to the privacy level.

\subsection{Performance Analysis}
\begin{theorem}[Performance of Algorithm 1]
	For an $N$-client distributed learning problem in Eq.~\eqref{optim_problem}, given the communication constraint $\bar{b}^{(i)}$, privacy level $(\bar{\epsilon}^{(i)}, \bar{\delta}^{(i)})$, clipping norm bound $C$, learning rate $\eta \le \frac{1}{\nu}$ and training iterating $T$, BQ-SGD satisfies the following. \\		
	\textbf{Privacy:}  BQ-SGD is $(\sqrt{2T\log \frac{1}{\bar{\delta}^{(i)}}}\bar{\epsilon}^{(i)}, T\bar{\delta}^{(i)})$- SGD privacy for client $i$;\\
	\textbf{Communication:} BQ-SGD incurs communication cost $Td\bar{b}^{(i)}$ for client $i$;\\
	\textbf{Convergence:} The convergence error for smooth objectives $F(\theta_t)$ is upper bounded by
	\begin{align}
 \begin{array}{llll}
	&~~~~\frac{1}{T}\sum_{t=0}^{T-1}\mathbb{E}[\|\nabla F(\theta_t)\|^2]\\
  &\le \underbrace{\frac{2[F(\theta_0)-F(\theta^*)]}{T\eta} + \frac{N \sigma^2 \sum_{i=1}^N p_i^2}{L}}_{\text{\rm Error of Distributed SGD}}  \nonumber\\
		&~~ + \underbrace{NdC^2\sum_{i=1}^N p_i^2V(m^{(i)},\frac{1}{2},s^{(i)})}_{\text{\rm Quantization Error}},  
   \end{array}
	\end{align}
	where $m$ and $s$ are determined by Eqs.~\eqref{eq:s} and \eqref{eq:m}.
	\label{Theorem 1}
\end{theorem}
The communication and privacy performance can be directly derived from the training procedure, and the full proof of the convergence error bound is shown in Appendix~\ref{pro:The1}. 

$\bullet$ \textbf{Communication}. At each iteration, client $i$ quantizes each element of the $d$-dimensional gradient from $b_{init}$ to $\bar{b}^{(i)}$ bits ($b_{init}$ is the number of bits of full-precision floating point, e.g., $b_{init}=32$ or $b_{init}=64$). We can achieve this communication cost by summing over $T$ iterations. Compared with vanilla distributed SGD, we can reduce $Td(b_{init}-\bar{b}^{(i)})$ bits communication overhead for client $i$. 

$\bullet$ \textbf{Privacy}. Using the following strong composition Theorem~\cite{dwork2014algorithmic}, client $i$ has a total $(\sqrt{T\log \frac{1}{\bar{\delta}^{(i)}}}\bar{\epsilon}^{(i)}, T\bar{\delta}^{(i)})$- SGD privacy after repeating $T$ times BQ operation.

\begin{lemma}[Strong Composition Theorem~\cite{dwork2014algorithmic}]\label{lemma2}
	For all $\epsilon$, $\delta$ $\delta'\ge 0$, the class of $(\epsilon, \delta)$- DP mechanisms satisfies $(\epsilon^T, T\delta+\delta')$-differential privacy under $T$-fold adaptive composition for:
	\begin{align}
		\epsilon^T = \sqrt{-2T\ln{\delta'}}\epsilon + T\epsilon (e^{\epsilon} - 1).
	\end{align}
	For small $\epsilon$, $e^{\epsilon} -1 \to 0$, we have $\epsilon^T=O(\sqrt{T}\epsilon)$.\end{lemma}

$\bullet$ \textbf{Trade-off among privacy-communication-convergence}
We then fix one of $\bar{\epsilon}$ and $\bar{b}$ to see how the change of the other affects the values of $m$ and $s$.

\textit{Fix privacy constraint $\bar{\epsilon}$}. As $\bar{R}$ is proportional to $\bar{\epsilon}$, we turn to analyze the effect of $\bar{b}$ with fixed $\bar{R}$. Using Eq.~\eqref{eq:m}, we rewrite the BQ noise variance $V(m,\frac{1}{2},s)$ as: 
\begin{equation}
	V(m,\frac{1}{2},s)  = \frac{1}{4{\bar{R}}^2}+ \frac{1}{6s^2}.	
\end{equation}
For a small communication budget $\bar{b}$, we have to quantize the raw gradient with lower-level $2^{\bar{b}} -1$, which leads to lower-level $s$ according to Eq.~\eqref{eq:s}. For fixed $\bar{R}$ and lower $s$, the variance tends to be large, indicating that small-level quantization leads to large performance degradation.

\textit{Fix communication constraint $\bar{b}$}. For high privacy requirements $\bar{R}$, again using Eq.~\eqref{eq:m} and $m+2s=2^{\bar{b}} -1$, we can obtain a high-level $m$ and low-level $s$. This demonstrates that the algorithm needs to quantize the gradient to a small level to bring more noise and prefers to add more binomial noise as the additional source to complete $\bar{R}$ to achieve a high privacy guarantee. Taking such values of $s$ and $m$, the variance function $V(m,\frac{1}{2},s)$ tends to be large, which means high-level privacy leads to a large performance decay.

%% file: Experiments.tex
\section{Experiments}
\label{sec:experiments}
In this section, we conduct experiments on MNIST and Fashion-MNIST to empirically validate our proposed BQ-SGD.  We use LeNet-5~\cite{lecun2015lenet} for MNIST, and AlexNet~\cite{krizhevsky2012imagenet} for the Fashion-MNIST for all clients. More experimental details are given in Table~\ref{tab:parameter}. 
\vspace{-1pt}
\begin{table}[h]
	\caption{Experiment Setting.}
	\label{tab:parameter}
	\vspace{-10pt}
	\begin{center}
		\begin{small}
				\begin{tabular}{lcccr}
					\hline
					Dataset & MINIST & Fashion-MNIST \\ 
					\hline
					Net & LeNet-5 & AlexNet \\ 
					Dim for SGD Privacy\footnotemark[1] & $d_{P}=3\times 10^3$  & $d_{P}=3\times 10^4$ \\ 
					Clipping Threshold & $C=0.0015$  & $C=0.003$ \\ 
					Learning Rate & 0.006  & 0.006 \\ 
					Batch Size    & 32 & 32  \\ 
					Number of Clients       & 4  & 4  \\
					Size of Local Datasets & 15000 & 15000\\
					Iterations  & 1000  & 3000  \\ 
					\hline
				\end{tabular}
		\end{small}
	\end{center}
	\vskip -0.1in
\end{table}
\footnotetext[1]{we use $d_{P} = {\| \nabla l\|_1}/{C}$ instead of $d$ to give an approximate value of the privacy level.}

\textbf{Performance of BQ-SGD for different privacy constraints.} We first fix the communication budget $\bar {b}$ and examine the performance of BQ-SGD under various privacy demands. We set $\bar{b}=8$ bits, $\bar{\delta} =1 \times 10^{-4}$ for MNIST and $b^* = 10$ bits, $\bar{\delta} =1 \times 10^{-4}$ for Fashion-MNIST. From Table~\ref{tab:BQprivacy}, our BQ-SGD achieves the accuracy of $94.20\%$, $96.73\%$ and $97.38\%$ on the MNIST datasets with  $\bar{\epsilon} = 1.72, 3.44, 8.72$. That is,  the model performance deteriorates with the decrease in $\bar{\epsilon}$, indicating stronger privacy requirements correspond to worse model performance. Similar results can be found on Fashion-MNIST, our BQ-SGD  achieves the accuracy of $79.11\%$, $84.16\%$ and $87.02\%$ with $\bar{\epsilon} = 86.22, 112.42, 138.79$. In addition, when faced high privacy demand, i.e., small $\bar{\epsilon}$, we have small $s$, i.e., low quantization level and large $m$, i.e., high Binomial noise level since they can incur more noise to meet the requirements of privacy, which is consistent with our conclusion in Theorem~\ref{Theorem 1}.


\vspace{-1pt}
\begin{table}[htbp]
	\caption{BQ-SGD with fixed $\bar{b}$ and different $\bar{\epsilon}$($\bar{\delta} =1 \times 10^{-4}$)}
    \label{tab:BQprivacy}
    \vspace{-10pt}
	\begin{center}
		\begin{small}
				\begin{tabular}{cccc}
					\hline
					Dataset    	&$\bar{\epsilon}$&  BQ parameters  &Accuracy  \\ \hline
					\multirow{3}{*}{\begin{tabular}[c]{@{}l@{}}MNIST\\ ($\bar{b} = 8$ bits )\end{tabular}}  
				    &  $1.72$ & $s=1, m=251$  & $94.20\%$ \\ 
					&  $3.44$ & $s = 2, m=251$    &  $96.73\%$ \\
					& $8.72$  &  $s = 4, m=247$  & $97.38\%$  \\ \hline
					\multirow{3}{*}{\begin{tabular}[c]{@{}l@{}}Fashion-MNIST\\ ($\bar{b} = 10$ bits)\end{tabular}}  
					&  $86.22$ & $s=10, m=1003$  & $79.11\%$  \\ 
					&  $112.42$ & $s = 13, m=997$    &  $84.16\%$ \\
					& $138.79$  &  $s = 16, m=991$  & $87.02\%$  \\ \hline
				\end{tabular} 
		\end{small}
	\end{center}
	\vskip -0.1in
\end{table}



\textbf{Performance of BQ-SGD for different communication budgets.} 
We next fix the privacy constraint and explore the effect on communication budget $\bar{b}$. We set $\bar{\epsilon} = 3.44$, $\bar {\delta} = 10^{-4}$ for MNIST and $\bar{\epsilon} = 112.42$, $\bar{\delta} = 1 \times 10^{-4}$ for Fashion-MNIST. Table~\ref{tab:BQcomm} demonstrates that as the communication budget increases, the performance of our BQ-SGD improves.
In particular, on MNIST, the BQ-SGD achieves the accuracy of $96.73\%$, $96.91\%$, and $96.99\%$ using $\bar{b} = 8, 10, 12$ bits. And on Fashion-MNIST, the BQ-SGD achieves the accuracy of $84.16\%$, $84.71\%$, and $85.02\%$ using $\bar{b} = 10, 12, 14$ bits Furthermore, it can be seen that both $s$ and $m$ grow when the communication constraints are relaxed. On the one hand, clients are allowed to use a high quantization level to ensure learning performance. On the other hand, a larger $s$ incur less disturbance to the raw gradient, thus a larger binomial noise needs to be added to ensure privacy.

\begin{table}[htbp]
	\caption{BQ-SGD with fixed $\bar{\epsilon}$ and different $\bar{b}$ ($\bar{\delta} =1 \times 10^{-4}$).}
	\label{tab:BQcomm}
	\vspace{-10pt}
	\begin{center}
		\begin{small}
\begin{tabular}{cccc}
	\hline
	Dataset   &$\bar{b}$(bits)&  BQ parameters  &Accuracy  \\ \hline
	\multirow{3}{*}{\begin{tabular}[c]{@{}l@{}}MNIST\\ ($\bar{\epsilon} = 3.44$ )\end{tabular}}      
	&  $8$ & $s=2, m=251$  & $96.73\%$  \\ 
	&  $10$ & $s = 4, m=1050$    &  $96.91\%$ \\ 
	& $12$  &  $s = 6, m=4079$  & $96.99\%$  \\ \hline
	\multirow{3}{*}{\begin{tabular}[c]{@{}l@{}}Fashion-MNIST\\ ($\bar{\epsilon} = 112.42$)\end{tabular}}  
	&  $10$ & $s = 13, m=997$  & $84.16\%$  \\
	&  $12$ & $s = 26, m=4043$    &  $84.71\%$ \\ 
	& $14$  &  $s = 52, m=16279$  & $85.02\%$  \\ \hline
\end{tabular} 
	\end{small}
	\end{center}
	\vskip -0.1in
\end{table}

%% file: Conclusion.tex
\section{Conclusion}
We provided new insights into the correlated nature of quantization and privacy. Using the inherent privacy properties, we proposed a new quantization-based solution to achieve communication efficiency and privacy protection simultaneously. We demonstrated the effectiveness of our proposed solutions in the distributed SGD framework and theoretically characterized the new trade-offs among communication, privacy, and learning performance. 

%% file: Appendix.tex
\section*{Appendix}

\subsection{Proof of Proposition 1}
\label{pro:Pro1}
\begin{proof}
	The noise is raised from two sources, one is the disturb caused by uniform quantization and another is from added Binomial noise. For simplicity, we omit $(i)$ and $t$ in the following analysis. We first decompose the BQ noise as $\bm{r}^B =  \bm{r}^U+ \frac{C}{s}\mathbf{o} - \frac{Cmq}{s}\cdot\bm{1}$, where $\bm{r}^U$ is incurred by the uniform quantizer and $\frac{C}{s}\mathbf{o} - \frac{Cmq}{s}\cdot\bm{1}$ is generated from Binomial noise.
	We then derive the analysis in two steps.
	
	\textbf{Step 1: Properties of Uniform quantization noise}
	We first show the statistic properties of the uniform quantization noise $\mathbf{r}^{U}$. Considering $\mathcal{Q}_b[g_j] = C \cdot \text{sgn}(g_j)\cdot\rho(g_j,s)$, then we define $r^U_j= g_j-\mathcal{Q}_b[g_j]$, i.e., the noise added on the clipped gradient after uniform quantization. 
	
	To analyze $r^U_j$, we introduce an intermediate variable $\varphi_j \triangleq \frac{|g_j|}{C} - \rho(g_j,s)$. Denote $\bar f_{g}(x) = \sum_{l=0}^{s-1} f_{g}(\cfrac{l}{s}+x), x\in[0,\frac{1}{s})$ as the PDF of ${|g|_j}/{C}$ in a single quantization bin. Then for $0 < \varphi_j < \frac{1}{s}$, the probability density of $\varphi_j$ is
	\begin{align*}
		f_{\varphi_j}\{\varphi_j\} &= \sum_{l=0}^{s-1} f_{g}\Big\{\frac{|g_j|}{C}=\frac{l}{s}+\varphi_j\Big\}\cdot \Delta  \\
		&~~~\times \Pr\Big\{\rho(g_j,s)=\frac{l}{s} \Big\vert \frac{|g_j|}{C}=\frac{l}{s}+\varphi_j\Big\}/ \Delta\\
		&= \sum_{l=0}^{s-1} f_{g}\Big\{\frac{|g_j|}{C}=\frac{l}{s}+\varphi_j\Big\} \cdot \frac{\frac{1}{s}-\varphi_j}{\frac{1}{s}}\\
		&\triangleq \bar f_{g}(\varphi_j)  [1 - s\varphi_j].
	\end{align*}	
	where $\bar f_{g}(\varphi_j) \triangleq \sum_{l=0}^{s-1} f_{g}\Big\{\frac{|g_j|}{C}=\frac{l}{s}+\varphi_j\Big\}$, and $\Delta$ is a small region containing the point $\frac{|g_j|}{C}=\frac{l}{s}+\varphi_j$. 
	
	Similarly, for $-\frac{1}{s}<\varphi_j<0$, we have:
	\begin{align*}
		f_{\varphi_j}\{\varphi_j\}  &= \sum_{l=0}^{s-1} f_{g}\Big\{\frac{|g_j|}{C}=\frac{l+1}{s}+\varphi_j\Big\}\cdot \Delta\\ &~~~~\times \Pr\Big\{\rho(g_j,s)=\frac{l+1}{s} \Big\vert \frac{|g_j|}{C}=\frac{l+1}{s}+\varphi_j\Big\}/ \Delta\\
		&= \sum_{l=0}^{s-1} f_{g}\Big\{\frac{|g_j|}{C}=\frac{l+1}{s}+\varphi_j\Big\} \cdot \frac{\frac{1}{s}+\varphi_j}{\frac{1}{s}}\\
		&\triangleq \bar f_{g}(\frac{1}{s}+\varphi_j) [1 + s\varphi_j].
	\end{align*}
    Note we have no prior knowledge of the distribution of $g_j$ due to the randomness of gradient calculation at each iteration. Such randomness comes from optimization algorithms (e.g., model initialization, batch selection) or hardware defaults. Based on the principle of indifference~\cite{eva2019principles}, the best that we can assume that the normalized gradient $\frac{|g_j|}{C}$ is uniformly distributed in each quantization bin. Under this assumption, we have $\bar f_{g}(x) = s$ and thus,
	\begin{equation}
		f_{\varphi_j}\{\varphi_j\} = s - s^2|\varphi_j|,~~~\varphi_j\in [-\frac{1}{s}, \frac{1}{s}]
	\end{equation}
 
	Considering that $r^U_j = C \cdot \text{sgn}(g_j)\cdot\varphi_j$ and $f_{\varphi_j}\{\varphi_j\}$ is even function, we have probability density of $r^U_j$ is 
	\begin{equation}
		f_{\mathbf{r}}(r^U_j) = \frac{s}{C} - \frac{s^2}{C^2}|r^U_j|,~~~r^U_j\in [-\frac{C}{s}, \frac{C}{s}]
	\end{equation} 
	
	\textbf{Step2: properties of BQ noise}
	Recall that $\bm{r}^B =  \bm{r}^U+ \frac{C}{s}\mathbf{o} - \frac{Cmq}{s}\cdot\bm{1}$. Define a temporary vector $\bm{\psi} = \bm{r}^U + \frac{C}{s}\mathbf{o}$. For $\psi_j \in [\frac{kC}{s}, \frac{(k+1)C}{s}]$, the probability density of $\psi_j$ is
	\begin{align*}
		&f_{\bm{\psi}}\{\psi_j\} \\
		&= \Big[f_{\mathbf{r}}\Big\{r^U_j = \psi_j - \frac{(k+1)C}{s}\Big\}  \cdot \Delta  \cdot P_{k+1} \\ 
		&~~~+ f_{\mathbf{r}}\{r^U_j = \psi_j - \frac{kC}{s}\}  \cdot \Delta  \cdot  P_k\Big]/\Delta\\
		&= \Big[\frac{s}{C} - \frac{s^2}{C^2}( \frac{(k+1)C}{s} -\psi_j)\Big]P_{k+1}\\
		&~~~+ \Big[\frac{s}{C} - \frac{s^2}{C^2}(\psi_j - \frac{kC}{s})\Big]P_k\\
		& = \frac{s}{C}[(k+1)P_k - kP_{k+1}] + \frac{s^2}{C^2}(P_{k+1}-P_k)\psi_j,
	\end{align*}
	for $k=-1,0,...,m$, where $\Delta$ is a small region for $r^U_j$, and $P_k = \binom{m}{k} q^k(1-q)^{m-k}$ is the probability value of $\mathbf{o}$ at $k$.
	We finally derive the probability density function of the BQ noise $r_j = \psi_j - \frac{Cmq}{s}$ as 
	\begin{align*}
		&f_{\bm{r}}\{r_j\}  \\
		&= \frac{s}{C}[(k+1)P_k - kP_{k+1}]+ \frac{s^2}{C^2}(P_{k+1}-P_k) (r_j + \frac{Cmq}{s})\\
		&= \frac{s[(k+1-mq)P_k + (mq-k)P_{k+1}]}{C}+ \frac{s^2(P_{k+1}-P_k) r_j}{C^2}
	\end{align*}
	for $r_j \in [\frac{(k-mq)C}{s}, \frac{(k+1-mq)C}{s}]$.
	
\end{proof}

\subsection{Proof of Lemma 1}
\label{pro:lemma1}

\begin{proof}
	We abbreviate Eq.~\eqref{eq:noise} as $f_{\bm{r}}(r_j)  \triangleq a_k + b_kr_j$. Let $\mathbf{g},\mathbf{g}'$ be clipped gradients and $\mathbf{g}^B,\mathbf{g}^{B'}$ be the corresponding BQ gradients. Using the definition of DP, we are looking at the difference for the same output $\mathbf{y}$,
	\begin{align*}
		& |\ln{\frac{\Pr(\mathbf{g}^{B'}=\mathbf{y})}{\Pr(\mathbf{g}^B=\mathbf{y})} |}=\Big|\ln{\Big\{\prod_{j=1}^d \frac{a_{k'_j} + b_{k'_j}(t_j - {g}'_j)}{a_{k_j} + b_{k_j}(t_j - {g}_j)}\Big\}}\Big|\\
		&\le \sum_{j=1}^d \Big| \ln{\Big\{\frac{a_{k'_j} + b_{k'_j}(t_j - {g}'_j)}{a_{k_j} + b_{k_j}(t_j - {g}_j)}\Big\}} \Big|\\
		&\overset{(a)}{\le} \sum_{j=1}^d \ln  \{1 + \frac{(|b_{k_j}|+|b_{k_j+1}|)\cdot|{g}'_j-{g}_j|}{a_{k_j} + b_{k_j}r_j} \},
	\end{align*}
	where $(a)$ considers ${g}_j$ and ${g}'_j$ are located at adjacent quantization bin (i.e., $|{g}'_j-{g}_j| < \frac{C}{s}$) if we take $L>2s$. (Without losing generality, suppose that $k'_j > k_j$). Let $z_j = \ln  \{1 + \frac{(|b_{k_j}|+|b_{k_j+1}|)\cdot|{g}'_j-{g}_j|}{a_{k_j} + b_{k_j}\zeta_j} \}$, then
	\begin{align}
		&\mathbb{E}[z_j] =\sum_{k_j=-1}^m \int_{\frac{(k_j-mq)C}{s}}^{ \frac{(k_j+1-mq)C}{s}} z_j \Big[a_{k_j} + b_{k_j}\zeta_j\Big] d\zeta_j\notag\\
		&\le \sum_{k_j=-1}^m \int_{\frac{(k_j-mq)C}{s}}^{ \frac{(k_j+1-mq)C}{s}} \frac{(|b_{k_j}|+|b_{k_j+1}|)\cdot|{g}'_j-{g}_j|}{a_{k_j} + b_{k_j}\zeta_j}\notag\\
		&~~~~~~~~~~~~~~~~~~~~~~~~~~~\times [a_{k_j} + b_{k_j}\zeta_j] d\zeta_j\notag\\
		&=\frac{C\cdot|{g}'_j-{g}_j|\sum_{k_j=-1}^m (|b_{k_j}|+|b_{k_j+1}|)}{s}\notag\\
		&=\frac{2s\cdot|{g}'_j-{g}_j|\sum_{k_j=-1}^m |P_{k_j+1}-P_{k_j}|}{C}\notag\\
		&\le \frac{4s\cdot|{g}'_j-{g}_j|P_{max}}{C}\notag,
	\end{align} 
	where $P_{max} = \max_k\{P_k\}$. Hence,
	\begin{align*}
		\mathbb{E}[\sum_{j=1}^d z_j] = \sum_{j=1}^d \mathbb{E}[z_j]\le\frac{4s\Delta\mathbf{g}P_{max}}{C}
	\end{align*}
	where $\Delta\mathbf{g} = \max_{D,D'}\|\mathbf{g}-\mathbf{g}'\|_1$ is the sensitivity function. According to the Markov's inequality, we have $\Pr\Big[\sum_{j=1}^d z_j \ge \epsilon \Big] \le \frac{4s\Delta\mathbf{g}P_{max}}{C\epsilon'} = \delta'$. Note that $\Delta\mathbf{g}$ gives an upper bound on how much we must perturb the output to preserve privacy, therefore it only relates to the norm clipping operation with threshold $C$. Using Eq.~(\ref{eq:clipgrad}), we can rewrite the sensitivity function as:
	\begin{equation}
		\begin{array}{llll}	
			\Delta\mathbf{g} = \frac{1}{L} \max_{\xi,\xi'}\|\nabla \hat{l}(\xi)-\nabla \hat{l}(\xi')\|_1 \le \frac{2dC}{L}\end{array},  \label{eq:sensitivity}
	\end{equation}
	where $\xi$ and $\xi'$ are the single entries which are different for datasets $D$ and $D'$. Leveraging this bound, we have $\epsilon' = \frac{8dsP_{max}}{L\delta'}$. In addition, $\tilde{\mathbf{g}}_t^{(i),B}$ is computed over batch $B_t^{(i)}$, which is randomly sampled from dataset $D^{(i)}$. therefore, according to the privacy amplification theorem~\cite{balle2018privacy}, $\tilde{\mathbf{g}}_t^{(i)}$ at client $i$ assures $(\frac{L}{|D^{(i)}|}\epsilon', \frac{L}{|D^{(i)}|}\delta')$-differential privacy. Hence, the BQ gradient satisfied $(\epsilon^{(i)}, \delta^{(i)})$-differential privacy for client $i$ in one communication round, where $\epsilon^{(i)} = \frac{8 dsLP_{max}}{|D^{(i)}|^2 \delta^{(i)}}.$
\end{proof}

\subsection{Proof of Theorem 1}
\label{pro:The1}
\begin{proof}
Combining Assumption~\ref{ass:stochastic_gradient} and Eq.~\eqref{eq:BQvariance} in Proposition 1, the properties of aggregated gradient ${\mathbf{\bar g}}_t$ at server satisfy:
\begin{align}
	&\mathbb{E}_{B_i,\mathcal{Q}}[{\mathbf{\bar g}}_t] = \nabla F(\theta_t),\label{eq:unbiassness}\\
	&\mathbb{E}_{B_i,\mathcal{Q}}\left[||{\mathbf{\bar g}}_t||^2\right] \le \|\nabla F(\mathbf{x}_t)\|^2 
	+ \frac{N\sigma^2\sum_{i=1}^N p_i^2}{L} \nonumber\\
     &~~~~~~~~~~~~~~~~~+ dNC^2\sum_{i=1}^N p_i^2V(m^{(i)}, q, s^{(i)}).
	\label{eq:qsg}
\end{align}

Firstly, we consider function $F$ is $\nu\text{-smooth}$, and use Eq.~\eqref{eq:smooth_1}:
\begin{equation}\nonumber
	F(\theta_{t+1}) \le F(\theta_t) + \nabla F(\theta_t)^\text{T} (\theta_{t+1}-\theta_t) + \frac{\nu}{2} \|\theta_{t+1}-\theta_t\|^2.
\end{equation}

For the BQ-SGD, $\theta_{t+1} = \theta_t -\eta \mathbf{\bar g}_t $, so:
\begin{align*}
	F(\theta_{t+1})&\le F(\theta_t) + \nabla F(\theta_t)^\text{T} (-\eta \mathbf{\bar g}_t) + \frac{\nu\eta^2}{2} \|\mathbf{\bar g}_t\|^2.
\end{align*}

Taking total expectations and using Eq.~\eqref{eq:unbiassness} and \eqref{eq:qsg}:
\begin{equation}\nonumber
	\begin{split}
		&\mathbb{E}[F(\theta_{t+1})] \le F(\theta_t) + (-\eta + \frac{\nu \eta^2}{2})\|\nabla F(\theta_t)\|^2 \\  
  &+ \frac{\nu\eta^2N\sigma^2\sum_{i=1}^N p_i^2}{2L} + \frac{\nu\eta^2dNC^2\sum_{i=1}^N p_i^2V(m^{(i)}, q, s^{(i)})}{2}.
	\end{split}
\end{equation}

Subtracting $F(\theta_t)$ from both sides, and for $\eta \le \frac{1}{\nu}$
\begin{align*}
	\mathbb{E}[F(\theta_{t+1})]- F(\theta_t)&\le -\frac{\eta}{2}\|\nabla F(\theta_t)\|^2 + \frac{N\eta\sigma^2\sum_{i=1}^N p_i^2}{2L}  \\
	& + \frac{N\eta dC^2\sum_{i=1}^N p_i^2V(m^{(i)}, q, s^{(i)})}{2}.
\end{align*}
Applying it recursively, this yields:
\begin{align*}
	\mathbb{E}[F(\theta_T)]-F(\theta_0) &\le -\frac{\eta}{2} \sum_{t=0}^{T-1} \|\nabla F(\theta_t)\|^2+ \frac{\eta N \sigma^2T\sum_{i=1}^N p_i^2}{2L} \\
	&+ \frac{\eta N dTC^2\sum_{i=1}^N p_i^2V(m^{(i)}, q, s^{(i)})}{2}.
\end{align*}
Considering that $F(\theta_T) \ge F(\theta^*)$, so:
\begin{align*}
	&\frac{1}{T}\sum_{t=0}^{T-1}\mathbb{E}[\|\nabla F(\theta_t)\|^2] \le \frac{2[F(\theta_0)-F(\theta^*)]}{T\eta}\\
	&~~ + \frac{N\sigma^2\sum_{i=1}^N p_i^2}{L} + N d C^2\sum_{i=1}^N p_i^2V(m^{(i)}, q, s^{(i)}).
\end{align*}
\end{proof}